# Heterogeneous Effects of Green Finance on Urban Decarbonization: Evidence from 285 Cities in China

**Author1: Xueyang Li ;Author2: Jinlei Ma**

**Affiliation:**

[1and2]School of Business, Anhui University of Technology, Anhui 243002, China

**All authors' email:**

[a]Email:lixueyang0618@163.com

[a]Email: mjl19905464288@163.com

**Abstract:** While green finance has become a key instrument for low-carbon city transitions, its actual decarbonization effects and transmission mechanisms remain unclear. This study employs econometric models and machine learning-based analysis to examine whether and how green finance reduces city-level carbon intensity. Results show that green finance significantly lowers carbon intensity, with green bonds and green investment having the strongest impacts and evident spatial spillovers. The effects vary by development level, being most pronounced in Fourth- and Fifth-tier cities. Mediation analysis reveals that green finance operates mainly through energy structure optimization, followed by industrial upgrading, foreign direct investment, and technological innovation. SHAP analysis confirms substantial differences across financial instruments, with green bonds, funds, and credit contributing most to decarbonization. Moreover, the marginal impact is stronger in cities with low technological capacity, high industrial dependency, and coal-based energy mixes. These findings provide theoretical support and policy guidance for building a multi-level, regionally differentiated green finance system to promote inclusive low-carbon transitions.

## 1. Introduction

Addressing climate change has become one of the most urgent global challenges in the 21st century[1]. The intensifying impacts of global warming, driven primarily by rising greenhouse gas emissions[2], are already manifesting in more frequent and severe extreme weather events[3], biodiversity loss[4], increasing risks to food[5] and water security[6]. According to the research of [7], the global average temperature in 2024 exceeded pre-industrial levels by 1.5°C for the first time, signaling a critical threshold outlined by the Paris Agreement. In this context, accelerating carbon emission reduction has become a pressing policy imperative for countries worldwide[8].

China, as the world's largest carbon emitter, plays a central role in global climate governance[9] and has pledged to peak carbon emissions before 2030 and achieve carbon neutrality by 2060[10]. However, achieving this dual-carbon target requires reconciling carbon reduction goals with economic development needs[11]. A key challenge lies in facilitating a decoupling of carbon emissions from economic growth, particularly in cities where energy consumption, industrial activity, and population density are highly concentrated[12, 13]. Given that city areas account for over 80% of the country's total carbon emissions[14], cities have emerged as crucial arenas for implementing climate mitigation strategies[15].

Green finance, serving as a crucial bridge between financial resources and environmental objectives, has increasingly become a key instrument in facilitating low-carbon transitions[16]. Existing studies suggest that green finance significantly affects the evolution of carbon intensity through dual channels of capital allocation and technological innovation[17-19]. Instruments such as green credit and green bonds channel financial resources toward green infrastructure, renewable energy projects, and environmental technology R&D, thereby contributing to emission reduction goals[20-22]. For example, several empirical studies employing panel regressions or quantile regressions have found that green credit or green bonds significantly reduce carbon intensity at the national or provincial level, particularly in regions with a stronger foundation for green industries[23-25]. Xu and Xu [26] have further shown that green investment can enhance R&D capacity in research institutions, promote breakthroughs in carbon capture and storage (CCS) technologies, and consequently facilitate deeper decarbonization. Moreover, some studies incorporating mediation analysis have revealed that green finance contributes to carbon reduction indirectly by enhancing city green innovation capabilities or facilitating industrial upgrading[27, 28]. In addition, provincial level empirical evidence also suggests that continuous improvement in green finance can enhance financial efficiency and accelerate the transition to low-carbon energy by supporting the development of renewable energy and alternative technologies[19].

However, several limitations remain in the existing literature. First, most studies have been conducted at the national[29], provincial[30, 31], or industry level[32, 33], neglecting the substantial heterogeneity that exists across cities in China in terms of economic development, financial accessibility, industrial structure, and policy enforcement capacity[34]. Even within the same province, cities differ considerably in the conditions under which green finance is implemented and in the pathways through which policies are transmitted. As such, macro-level analyses are often insufficient for capturing the micro-level mechanisms through which green finance influences city carbon emissions. Second, the majority of studies rely on linear econometric frameworks, which fail to account for the nonlinear marginal effects of green finance across cities at different stages of development[29, 35]. For instance, green finance may exhibit limited decarbonization effects at an early stage, but stronger effects once policy maturity and financial infrastructure reach critical thresholds[35, 36]. The assumption of homogeneous average decarbonization effects under linear models tends to obscure the dynamic and context-dependent nature of policy effectiveness.

In addition, current research remains relatively weak in identifying the spatial externalities and structural heterogeneity associated with green finance. On the one hand, the development of green finance in one city is not only shaped by local conditions, but also influenced by neighboring cities through channels such as policy demonstration, capital flows, and technological diffusion[28, 31]. Cities that are pioneers in green finance may exert spillover effects on surrounding areas through demonstration-emulation mechanisms, facilitating regional low-carbon coordination[37]. Nevertheless, most existing studies overlook such inter-city spatial dependencies, and fail to explicitly capture the cross-city spillover effects and spatial transmission mechanisms of green finance. On the other hand, cities vary greatly in terms of financial resource availability, institutional settings, industrial capacity, and green technological capability. These structural differences strongly influence the transmission pathways and effectiveness of green financial policies, but few studies have constructed frameworks capable of identifying nonlinear marginal effects or threshold conditions. Although some studies have attempted to incorporate regional groupings or heterogeneity analysis, most remain limited to crude categorizations such as "Eastern-Central-Western" or city size divisions[28, 38]. As a result, fundamental questions, such as "In what types of cities is green finance most effective?" and "At which stages of development do marginal returns peak?", remain insufficiently explored.

To address these gaps, this study constructs an integrated empirical identification framework that combines econometric modeling with machine learning techniques, using panel data of 285 prefecture-level cities in China from 2000 to 2022. By jointly applying fixed effects models, spatial econometric approaches, and causal forest algorithms, this study systematically evaluates the carbon-reduction effects of green finance at the city scale. Specifically, this article focuses on the following three core issues: (1) Has green finance significantly reduced carbon emission intensity at the city level? (2) Is there a significant difference in the decarbonization effects across cities at different stages of development? And (3) does the marginal role of green finance exhibit heterogeneity due to the development stage, industrial structure, technological capabilities, and other conditions of the city, resulting in non-linear changes in the decarbonization effect during the city development process?

Compared with existing studies, this research makes three key contributions. First, it introduces a novel analytical focus by exploring the average, spillover, heterogeneous, and nonlinear effects of green finance on decarbonization, which remain unexplored in the literature. Second, it shifts the scale of analysis to the city level, highlighting cities as critical yet understudied units in climate governance and enabling finer-grained identification of green finance impacts across diverse local contexts. Third, it employs a multi-method empirical framework that random panel model, spatial econometric models, and causal forests model, enhancing causal inference while capturing spatial and structural heterogeneity. These contributions together offer new insights into how green finance supports low-carbon transitions and inform more targeted city policy under China's dual-carbon strategy.

The rest of this paper is organized as follows: Section 2 proposes theoretical hypotheses; Section 3 details the methodology and data; Section 4 presents and discusses the research results; Section 5 summarizes the study and offers policy recommendations.

## 2. Theoretical hypotheses

### 2.1 Direct effects of green finance on carbon intensity

Theoretically, green finance constitutes a critical policy tool for facilitating the transition toward low-carbon development by reallocating financial resources toward environmentally sustainable sectors [39, 40]. Drawing on the theoretical framework of environmental Kuznets curve (EKC) and resource allocation efficiency, green finance may reduce urban carbon intensity through several channels. First, by alleviating credit constraints for green enterprises, green finance enables greater capital inflows to clean energy, energy-saving retrofits, and low-carbon infrastructure [41]. Second, through instruments such as green credit and green bonds, it reduces the cost of capital for environmentally beneficial projects, thereby enhancing their relative attractiveness carbon-intensive alternatives [42]. Third, green insurance and fiscal green funds can internalize environmental risks and promote technological upgrading by creating price signals and performance-based incentives [43].

However, the effectiveness of green finance in reducing carbon emission intensity is not universally assured. On the one hand, the allocation of green financial instruments may suffer from issues of greenwashing, regulatory arbitrage, or policy misalignment[44]. On the other hand, the actual emission-reduction effects may vary depending on the type, scope, and governance mechanisms of the green finance instruments employed. Therefore, it is important to examine whether the theoretical expectations of green finance translate into statistically verifiable reductions in carbon intensity, especially at the city level where implementation and enforcement vary widely.

Moreover, recent studies suggest that the impact of green finance on carbon emission intensity may operate through several key mediating mechanisms. First, green finance can promote green technological innovation by alleviating financing constraints for low-carbon enterprises and increasing investment in clean technology R&D. This enhances urban decarbonization capacity through the diffusion of energy-saving and renewable technologies [45, 46]. Second, green finance can influence the structure of energy consumption by directing capital flows toward clean energy and away from fossil fuel–intensive sectors, thereby improving the overall energy efficiency of urban economies [47, 48]. Third, green finance contributes to industrial structure upgrading by incentivizing the development of low-carbon, high-value-added industries, while restricting polluting and energy-intensive production through differentiated credit allocation and green bond standards [49-51]. Fourth, green finance affects carbon intensity by influencing the composition of foreign direct investment (FDI). Cities with stronger green financial systems tend to attract cleaner FDI with advanced technologies, while simultaneously deterring high-emission projects through stricter environmental financing standards. This shift improves the environmental quality of capital inflows and reduces local carbon intensity [52-55]. These four pathways—green innovation, energy structure, industrial upgrading, and FDI - jointly shape the mediating role of green finance in achieving low-carbon urban development.

Prior research has offered mixed evidence. Macro-level studies based on national or provincial data often find a negative association between green finance development and carbon emissions [56-58], but these studies typically rely on linear panel models and average effects, with limited causal identification and little attention to intranational heterogeneity. In contrast, city-level analyses, which are crucial given cities' central role in carbon governance and spatial variation in economic and regulatory contexts, remain scarce and fragmented.

**Hypothesis 1a:** Green finance exerts a statistically significant negative impact on city-level carbon emission intensity.

**Hypothesis 1b:** Different components of green finance (e.g., green credit, green bonds, green insurance) exert heterogeneous effects on urban carbon intensity.

**Hypothesis 1c:** Green finance reduces urban carbon intensity through multiple mediating channels, including green innovation, energy structure optimization, industrial upgrading, and the quality of foreign direct investment inflows.

**Hypothesis 1d:** The impact of green finance on carbon intensity varies across cities with different levels of socioeconomic development, with stronger effects in less developed cities.

### 2.2 Marginal nonlinear effects of green finance on decarbonization performance

The impact of green finance on urban decarbonization is unlikely to be linear or homogeneous across cities. Drawing on the principles of marginal utility theory and policy responsiveness heterogeneity [59, 60], the marginal effectiveness of green finance is expected to exhibit nonlinear patterns depending on local structural and institutional conditions. Specifically, the marginal benefit of green finance tends to be higher in cities where financing gaps, institutional inertia, or environmental inefficiencies are more severe, and lower where green investment mechanisms and infrastructure are already mature.

From a threshold effect perspective [61], cities undergoing structural transformation, such as mid-sized or resource-dependent localities, may display stronger responsiveness to green finance interventions due to more urgent demands for financial support, lower baseline performance in environmental governance, and higher marginal returns on green capital inputs. In contrast, in megacities with mature financial markets and highly developed green industries, green finance may face diminishing marginal effects as key sectors have already benefited from previous rounds of investment and institutional learning.

Moreover, heterogeneity in energy structures, technological capacity, and industrial composition introduces complex nonlinearities in the effectiveness of green finance. For example, green finance may have a stronger decarbonization impact in cities with moderate coal dependence, where substitution toward cleaner energy is feasible but not yet complete. In contrast, in cities with extreme fossil fuel reliance, deep-rooted institutional lock-ins and path dependency may attenuate the policy effect. Similarly, cities with limited green innovation capacity may benefit more from green finance as it serves a substitutional role in introducing external technologies and capital, while high-tech cities may already approach diminishing returns due to saturation effects [62].

**Hypothesis 2a:** The decarbonization effect of green finance exhibits nonlinear marginal patterns, with stronger effects at intermediate levels of coal dependence, industrial transition, or technological capacity.

**Hypothesis 2b:** The marginal effectiveness of green finance is more pronounced in structurally lagging or transitioning cities, compared to already green-intensive or financially mature cities.

## 3. Methodology

### 3.1 Methods

To comprehensively identify the impact pathways and functional characteristics of green finance on city carbon intensity, this study develops a multi-layered empirical strategy that integrates a random effects panel model, spatial econometric models, and a causal forest model. These models are respectively employed to identify causal effects, examine spatial spillover mechanisms, and explore heterogeneous impact patterns.

#### 3.1.1 Econometric model

First, based on panel data from prefecture-level cities, a random effects model is constructed to estimate the average impact of green finance on carbon intensity. The model is specified as follows:

$$CI_{it} = \alpha + \beta GF_{it} + \gamma Z_{it} + u_i + \varepsilon_{it} \tag{1}$$

where $CI_{it}$ denotes the carbon intensity of city $i$ in year $t$ ; $GF_{it}$ represents the level of green finance development; $Z_{it}$ is a vector of control variables, including industrial structure (IS), energy structure (ES), green technology (GT), economic growth (EG), foreign direct investment (FDI) and population density (PD); $u_i$ is the city fixed effect; $\varepsilon_{it}$ is the random error term.

To address potential endogeneity arising from reverse causality and omitted variable bias between green finance and carbon emissions, this study further employs a two-stage least squares (2SLS) approach. In the first stage, a set of instrumental variables is used, including lagged values of green finance (L.GF, L2.GF), the provincial average of green finance and its lags (AGF, L.AGF, L2.AGF), the governmental environmental attention and its lags (GEA, L.GEA, L2.GEA), as well as an indicator for policy pilot zones (Pilot). The first-stage regression is specified as follows:

$$GF_{it} = \theta_0 + \theta_1 IV_{it} + \theta_2 Z_{it} + \mu_i + \lambda_t + \varepsilon_{it} \tag{2}$$

In the second stage, the fitted values of green finance obtained from the first-stage regression $\widehat{GF}_{it}$ are used to replace the original explanatory variable, and the impact of green finance on carbon intensity is re-estimated accordingly.

$$CI_{it} = \alpha' + \beta' \widehat{GF}_{it} + \gamma' Zit + \mu_i + \lambda_t + \varepsilon_{it} \tag{3}$$

Furthermore, given that green finance may generate spatial spillover effects through mechanisms such as inter-city financial resource allocation, diffusion of green projects, or policy imitation, this study further employs spatial regression models to examine this mechanism.

$$CI_{it} = \beta_0 + \beta_1 GF_{it} + \beta_2 Z_{it} + \beta_3 WGF_{it} + \beta_4 WZ_{it} + \varepsilon_{it} \tag{4}$$

The spatial weight matrix $W$ is constructed based on an economically weighted geographic distance, which incorporates both geographical proximity and economic scale to capture inter-city linkages in green policy coordination and resource coupling. Lagged variables are generated through annual weighted averages aligned by intra-year city rankings, ensuring consistency between the structure of the matrix and the ordering of the data. The final model is estimated using ordinary least squares (OLS), with standard errors clustered at the city level to account for intra-group correlation in the panel data.

In constructing the spatial econometric model, this study adopts a nested spatial weight matrix $W$ that incorporates both geographic proximity and economic similarity. The economic distance matrix $W^{\text{econ}}$ is derived from the average per capita GDP (Per-GDP) of each city, where each element $W_{ij}^{\text{econ}}$ represents the inverse of the Euclidean distance in economic terms between cities $i$ and $j$. Similarly, the geographic distance matrix $W^{\text{geo}}$ is calculated based on longitude and latitude coordinates, with each element $W_{ij}^{\text{geo}}$ representing the inverse of the geographic distance. The initial nested weight matrix is constructed as the element-wise product of the two, i.e., $W_{ij} = W_{ij}^{\text{econ}} \cdot W_{ij}^{\text{geo}}$. To ensure conformity with spatial econometric modeling standards, the matrix is further normalized by (i) setting diagonal elements $W_{ii}$ to zero to exclude self-influence, and (ii) applying row-standardization so that $W_{ij}$ for all $i$. The resulting matrix is used as the normalized spatial weight matrix in subsequent estimation.

To uncover the underlying mechanisms through which green finance influences urban carbon emission intensity, this study adopts a causal mediation analysis framework. We quantify the extent to which the effects of green finance are transmitted through four key mediators: technological innovation, industrial structure, energy structure, and foreign direct investment (FDI). Following the framework of [63] and implemented via the medeff module by Hicks and Tingley [64], this approach enables the decomposition of total effects into direct and indirect components, thereby identifying the specific pathways through which green finance facilitates carbon reduction.

The mediation estimation proceeds in two stages. In the first stage, we regress the mediator $M_{it}$ (e.g., technological innovation) on green finance and control variables to assess the extent to which green finance influences the mediator. In the second stage, we regress carbon emission intensity $C_{it}$ on green finance, the mediator, and the same set of controls. The estimation equations are given as:

$$M_{it} = \theta_0 + \theta_1 GF_{it} + \theta_2 X_{it} + \mu_i + \lambda_t + \varepsilon_{it} \tag{5}$$

$$C_{it} = \delta_0 + \delta_1 GF_{it} + \delta_2 M_{it} + \delta_3 X_{it} + \mu'_i + \lambda' t + \varepsilon' it \tag{6}$$

where, $GF_{it}$ denotes the level of green finance in city i at time t; $M_{it}$ represents the mediating variable, which may be green technological innovation, industrial structure, energy structure, or foreign direct investment (FDI); and $C_{it}$ indicates the carbon emission intensity of city i. The control vector $X_{it}$ includes economic growth, population density, and environmental regulation. In addition, $\mu_i$ and $\lambda_t$ denote city and time fixed effects, respectively, while $\varepsilon_{it}$ and $\varepsilon'_{it}$ are idiosyncratic error terms.

### 3.1.2 XGBoost-SHAP model

While the baseline regressions and mechanism tests provide robust evidence on the causal relationship between green finance and urban carbon emission intensity, they may not fully capture the complex, nonlinear interactions among multiple influencing factors. In particular, traditional econometric models are limited in ranking the relative importance of predictors, especially when the effects are intertwined or potentially non-monotonic.

To further investigate the relative importance of green finance and other contextual factors in shaping urban carbon emission intensity, we extend the previous econometric analyses by incorporating a machine learning - based approach. Specifically, this study employed the eXtreme Gradient Boosting (XGBoost) model, a widely used ensemble learning algorithm known for its high accuracy and computational efficiency. The predictive model is trained using a supervised learning framework. Given a training set:

$$D = \{(x_i, y_i)\} \mathrm{i} = 1^n, \quad x_i \in R^p, \quad y_i \in R \tag{7}$$

where $x_i$ denotes the input feature vector and $y_i$ is the target variable, XGBoost predicts the output by aggregating K additive functions $f_k \in \mathcal{F}$ as

$$\hat{y}i = \sum k = 1^K f_k\left(x_i\right), \tag{8}$$

where each $f_k$ is a regression tree. The model is trained by minimizing the regularized objective:

$$L(\phi) = \sum i = 1^n 1\left(y_i, \hat{y}i\right) + \sum k = 1^K \Omega\left(f_k\right) \tag{9}$$

where l is the loss function (here, the squared error), and $\Omega\left(f_k\right) = \gamma T + \frac{1}{2}\lambda \mid \mathrm{w} \mid^2$ is the regularization term controlling model complexity, with T being the number of leaves and w the leaf weights.

To ensure model generalization and avoid overfitting, the input features were standardized using z-score normalization. The predictor set included eight core driving factors - GF, GT, EG, IS, ES, GEA, FDI, and PD - along with a temporal feature Year. The final model configuration was obtained via grid search and manual tuning: max depth = 6, learning rate = 0.1, n estimators = 900, and tree method = 'hist'.

Interpreting the contribution of each input factor to the model predictions is achieved through SHapley Additive exPlanations (SHAP), a unified interpretability framework grounded in cooperative game theory [65]. SHAP assigns each feature a Shapley value, representing its average marginal contribution to the model output across all possible feature coalitions. The prediction for a simplified input $z' \in \{0,1\}^M$ is represented as:

$$g\left(z'\right) = \phi_0 + \sum_{j=1}^{M} \phi_j z'_j \tag{10}$$

where $\phi_0$ denotes the model's baseline prediction, and $\phi_j$ represents the Shapley value of feature j, capturing its localized contribution to the prediction.

### 3.1.3 Causal forest model

To account for potential heterogeneity in the effects of green finance policies under varying city-level conditions, this study adopts a causal forest model based on the double machine learning (DML) framework to estimate the individualized treatment effects of green finance on carbon intensity. Let $Y$ denote carbon intensity, $T$ the level of green finance, and $X$ the set of covariates, including industrial structure (SIS), governmental environmental attention (GEA), and green technology level (GT). The objective of the model is to estimate:

$$\tau(x) = \mathbb{E}\left[Y(1) - Y(0) \mid X = x\right] \tag{11}$$

where $\tau(x)$ represents the marginal decarbonization effect of green finance under specific city-level characteristics. The model estimates the conditional average treatment effect (CATE) by separately modeling $Y$ and $T$ using random forests with varying parameter configurations and enhances estimation robustness through sample splitting and cross-fitting procedures.

### 3.2 Data and sources

#### 3.2.1 Variable definitions

(1) The dependent variable: carbon emission intensity, defined as the ratio of a city's total annual carbon emissions to its GDP for that year. This indicator measures the carbon emission level corresponding to a unit of economic output and is an important way to evaluate carbon reduction performance. Carbon emissions data sourced from Emissions Database Global Atmosphere Research (EDGAR-v2024)

(2) Core explanatory variable: development level of green finance (GF). This article uses the research results of Liu and He [66] to construct a green finance index for prefecture level cities. The index calculation is based on entropy method and comprehensively considers seven sub dimensions of green finance: Green Credit (proportion of environmental credit), Green Investment (intensity of pollution control investment), Green Insurance (penetration rate of environmental liability insurance), Green Bonds (proportion of green bonds), Green Support (proportion of fiscal environmental protection expenditure), Green Fund (proportion of green fund market value), and Green Equity (proportion of green equity transactions such as carbon trading).

(3) Control variables: the control variables mainly include industrial structure (proportion of secondary industry, IS), energy structure (proportion of coal consumption, ES), green technology innovation (number of green patent authorizations, GT), foreign direct investment (FDI), economic development (measured by the annual growth rate of GDP, EG), and population density (PD). These variables are incorporated to control for the influence of macroeconomic dynamics, structural conditions, institutional factors, and technological progress on carbon emission intensity.

(4) Instrumental variable: this study mainly uses governmental environmental attention (GEA) as an instrumental variable for green finance. GEA is measured by the frequency of environment - related keywords (e.g., such as “ecology” “low-carbon” “green” and “pollution”) in city-level government work reports, following Chen and Chen [67]. Government environmental attention can influence green finance development by guiding policy agendas and allocating resources toward environmentally oriented projects. At the same time, GEA is largely shaped by broader policy planning and public discourse rather than short- term carbon emission fluctuations [68, 69]. Although severe pollution events might at first glance pose an endogeneity concern - raising both emissions and governmental attention - government work reports are typically prepared annually, emphasizing long- term strategic priorities [70], which mitigates concern about reverse causality. Therefore, GEA satisfies both the relevance and plausibly the exogeneity criteria for a valid instrument.

Study also adopts the National Green Finance Reform and Innovation Pilot Zones (Pilot) as the second main instrumental variable. These zones, officially designated by the central government since 2017, aim to experiment with institutional innovations that facilitate green investment, financial product development, and credit support for environmental projects. Studies have shown that the selection of pilot cities was largely driven by administrative considerations such as regional balance, institutional capacity, and willingness to experiment with green financial reforms - rather than by cities' short-term environmental performance or carbon emissions [71, 72]. These findings suggest that the assignment of Pilot status introduces exogenous variation in green finance development, satisfying the key condition for instrument validity. Moreover, recent empirical analyses have adopted similar policy-based instruments in evaluating the effects of environmental and financial reforms, further reinforcing its theoretical soundness [73, 74].

**3.2.2 Data sources**

This study constructs a panel dataset of 285 prefecture level cities in China from 2000 to 2022. The data used are mainly from China City Statistical Yearbook, China Environmental Statistical Yearbook, China Financial Statistical Yearbook, and statistical yearbooks of provinces, cities, and autonomous regions. Some indicators are supplemented from the official websites of the National Bureau of Statistics, the China National Intellectual Property Administration, Wind database, city government work reports and other open channels.

Table 1 presents the descriptive statistics of the main variables. The standard deviation of carbon intensity (CI) is 2.96, indicating substantial variation in carbon emission performance across cities. Similarly, the green finance index (GF) and its sub-components also display notable dispersion, suggesting large differences in local green financial development. These heterogeneity highlights the need to investigate not only the average effect of green finance, but also its spatial spillovers and nonlinear marginal impacts across different structural contexts.

**TABLE 1 Summary statistics of variables**

| | Obs. | Mean | Std. | Min | Max |
|---|---|---|---|---|---|
| GF | 6555 | 28.50 | 10.83 | 4.07 | 65.75 |
| CI | 6555 | 2.88 | 2.96 | 0.07 | 33.21 |
| IS | 6555 | 45.89 | 11.40 | 10.68 | 90.97 |
| ES | 6555 | 0.70 | 0.33 | 0.00 | 4.44 |
| GT | 6555 | 246.13 | 942.55 | 0.00 | 18959.00 |
| EG | 6555 | 0.12 | 0.11 | -0.74 | 3.93 |
| FDI | 6017 | 45.38 | 112.81 | 0.00 | 1642.21 |
| PD | 6555 | 424.05 | 331.94 | 5.00 | 2712.00 |
| G-Credit | 6555 | 4.26 | 1.82 | 0.44 | 11.13 |
| G-Investment | 6555 | 1.06 | 0.47 | 0.11 | 2.98 |
| G-Insurance | 6555 | 1.91 | 0.83 | 0.21 | 5.31 |
| G-Bond | 6555 | 0.64 | 0.30 | 0.05 | 2.04 |
| G-Support | 6555 | 0.64 | 0.36 | 0.01 | 2.56 |
| G-Fond | 6555 | 4.28 | 1.84 | 0.43 | 12.40 |
| G-Equity | 6555 | 2.14 | 1.06 | 0.18 | 8.16 |

## 4. Results and discussion

### 4.1 Decarbonization effects of green finance

#### 4.1.1 Baseline regression results

To assess the average impact of green finance on city carbon intensity, a fixed effects model is employed for baseline estimation. The results are presented in Table 2. Column (1) shows that the coefficient of GF is -0.1575 and statistically significant at the 1% level, indicating that the development of green finance effectively reduces carbon intensity. Furthermore, Columns (2) to (8) in Table 2 demonstrate that various green financial instruments, including Green Credit, Green Investment, Green Insurance, Green Bonds, Green Support, Green Funds, and Green Equity, exhibit statistically significant negative effects on CI at the 1% level. These findings align with the existing literature suggesting that green finance can promote low-carbon development[14].

**TABLE 2 Results of baseline regression.**

| Dependent variable: CI | | | | | | | | |
|---|---|---|---|---|---|---|---|---|
| | (1) | (2) | (3) | (4) | (5) | (6) | (7) | (8) |
| GF | -0.1575*** | | | | | | | |
| | (-40.525) | | | | | | | |
| G-Credit | | -0.5213*** | | | | | | |
| | | (-30.950) | | | | | | |
| G-Investment | | | -1.7298*** | | | | | |
| | | | (-28.019) | | | | | |
| G-Insurance | | | | -1.0812*** | | | | |
| | | | | (-29.533) | | | | |
| G-Bond | | | | | -2.3169*** | | | |
| | | | | | (-24.232) | | | |
| G-Support | | | | | | -1.3186*** | | |
| | | | | | | (-19.882) | | |
| G-Fond | | | | | | | -0.5396*** | |
| | | | | | | | (-30.636) | |
| G-Equity | | | | | | | | -0.5885*** |
| | | | | | | | | (-23.173) |
| IS | -0.0363*** | -0.0321*** | -0.0325*** | -0.0321*** | -0.0311*** | -0.0302*** | -0.0328*** | -0.0323*** |
| | (-10.426) | (-8.5484) | (-8.5543) | (-8.5077) | (-8.0977) | (-7.7194) | (-8.7381) | (-8.2634) |
| ES | -1.1131*** | -1.1653*** | -1.1716*** | -1.1622*** | -1.1878*** | -1.2026*** | -1.1953*** | -1.1750*** |
| | (-9.6274) | (-10.085) | (-10.236) | (-10.153) | (-9.8144) | (-10.052) | (-10.815) | (-10.029) |
| GT | 0.0002*** | 0.0000 | -0.0000 | 0.0000 | -0.0000 | -0.0000 | 0.0000 | -0.0000 |
| | (8.8546) | (1.3383) | (-0.1384) | (0.8048) | (-0.3913) | (-1.2541) | (1.4601) | (-0.8187) |
| EG | 1.3540*** | 2.8338*** | 3.1408*** | 2.9030*** | 3.3538*** | 3.6971*** | 2.8572*** | 3.5216*** |
| | (5.4402) | (10.590) | (11.727) | (10.800) | (12.359) | (13.079) | (10.784) | (12.848) |
| FDI | -0.0011*** | -0.0021*** | -0.0022*** | -0.0022*** | -0.0025*** | -0.0030*** | -0.0022*** | -0.0026*** |
| | (-4.6414) | (-7.1343) | (-6.9929) | (-7.2657) | (-7.5992) | (-8.1688) | (-6.9238) | (-7.56403) |
| PD | -0.0002 | -0.0015*** | -0.0016*** | -0.0015*** | -0.0018*** | -0.0021*** | -0.0012*** | -0.0017*** |
| | (-0.7657) | (-3.7599) | (-3.6994) | (-3.6238) | (-3.8099) | ( -3.4365) | (-2.9159) | (-3.0926) |

Notes: t statistics in parentheses; ***, **, and * denote statistical significance at the 1%, 5%, and 10% levels, respectively.

Notably, the absolute magnitudes of the coefficients for these instruments differ substantially, even under the same level of statistical significance. This suggests that heterogeneous green financial instruments[1] vary in their effectiveness in reducing carbon intensity (Table 3). For example,

[1] All green finance sub-indicators are ratio-based and standardized across cities. The composite index is constructed using the entropy method, ensuring comparability of coefficients across different components.

G-Bond (β = -2.3169) and G-Investment (β = -1.7298) display larger absolute coefficients (Table 2), implying a stronger suppressive effect on carbon emissions, highlighting the central role of such green financial instruments in achieving decarbonization targets, and also indicating the critical effectiveness of fiscal guidance and policy intervention in the low-carbon transformation process[20]. In contrast, instruments such as G-Credit (β = -0.5213) and G-Fond (β = -0.5396) also exhibit robust decarbonization effects, but with smaller absolute coefficients. This may reflect differences in capital scale, investment focus, or time horizon, leading to variation in marginal emission reduction outcomes.

**TABLE 3 Results of Inter-group Statistical Tests.**

| Var1 | Var2 | Coefficient Difference | Standard Error | t-Statistic |
|---|---|---|---|---|
| G-Credit | G-Investment | 0.82*** | 0.08 | 10.12 |
| G-Bond | G-Equity | -1.21*** | 0.13 | -9.64 |
| G-Credit | G-Bond | 1.35*** | 0.12 | 11.00 |
| G-Bond | G-Fond | -1.34*** | 0.12 | -10.98 |
| G-Investment | G-Fond | -0.81*** | 0.08 | -10.08 |
| G-Investment | G-Equity | -0.68*** | 0.09 | -7.98 |
| G-Insurance | G-Bond | 0.98*** | 0.13 | 7.60 |
| G-Insurance | G-Fond | 0.37*** | 0.05 | 7.43 |
| G-Credit | G-Insurance | -0.36*** | 0.05 | -7.38 |
| G-Support | G-Fond | 0.73*** | 0.10 | 7.14 |
| G-Credit | G-Support | -0.73*** | 0.10 | -7.11 |
| G-Support | G-Equity | -0.59*** | 0.11 | -5.60 |
| G-Investment | G-Insurance | -0.45*** | 0.09 | -4.99 |
| G-Bond | G-Support | -0.23*** | 0.06 | -4.08 |
| G-Insurance | G-Equity | -0.62*** | 0.16 | -3.93 |
| G-Investment | G-Bond | 0.53*** | 0.14 | 3.68 |
| G-Fond | G-Equity | 0.14*** | 0.04 | 3.41 |
| G-Credit | G-Equity | 0.13*** | 0.04 | 3.33 |
| G-Insurance | G-Support | 0.36*** | 0.11 | 3.30 |
| G-Investment | G-Support | -0.09 | 0.13 | -0.69 |
| G-Credit | G-Fond | 0.00 | 0.03 | 0.11 |

#### 4.1.2 Instrumental variable regression results

In identifying the causal effect of green finance (GF) on city carbon emissions, potential endogeneity arising from reverse causality and omitted variable bias poses a critical concern. To mitigate this issue, we adopt an instrumental variable (IV) approach, drawing on the strategy of Xu, Fan [48], who addressed endogeneity in a similar context of green finance and carbon emission performance, and employ the following instruments: lagged terms of green finance (L.GF, L2.GF), provincial-level averages of green finance and their lags (AGF, L.AGF, L2.AGF), governmental environmental attention and its lags (GEA, L.GEA, L2.GEA), and an indicator for designation as a national green finance reform pilot zone (Pilot). These instruments are strongly correlated with city-level green finance but plausibly exogenous to the outcome variable. Specifically, the lagged values of GF reflect the cumulative influence of past policy and market dynamics on current green finance development; provincial averages capture decarbonization effects within regional financial systems; governmental environmental attention reflects local governments' prioritization of environmental issues in policy discourse, which shapes expectations and guides resource allocation, thereby indirectly stimulating green finance development; and the designation of national pilot zones serves as a formal policy intervention that enhances local governments' capacity and incentives to promote green finance[75]. These variables are theoretically justified, exhibit weak correlation with control variables (Table 4), thereby satisfying the exclusion restriction.

**TABLE 4 Correlation coefficients between instrumental variables and control variables.**

| Inspection tools: Pearson's r | | | | | | |
|---|---|---|---|---|---|---|
| Instrumental variable | IS | GT | ES | EG | FDI | PD |
| L.GF | -0.16 | 0.30 | -0.29 | -0.24 | 0.25 | 0.22 |
| L2.GF | -0.19 | 0.30 | -0.30 | -0.26 | 0.24 | 0.22 |
| AGF | -0.13 | 0.27 | -0.29 | -0.23 | 0.22 | 0.18 |
| L.AGF | -0.15 | 0.27 | -0.29 | -0.25 | 0.21 | 0.18 |
| L2.AGF | -0.17 | 0.27 | -0.30 | -0.27 | 0.20 | 0.18 |
| GEA | -0.02 | 0.18 | 0.01 | -0.16 | 0.15 | -0.01 |
| L.GEA | -0.05 | 0.18 | 0.00 | -0.20 | 0.15 | 0.00 |
| L2.GEA | -0.07 | 0.20 | -0.02 | -0.22 | 0.16 | 0.00 |
| Pilot | -0.02 | 0.11 | -0.02 | -0.03 | 0.06 | 0.01 |

Table 5 reports the first-stage regression results. All instruments are significantly associated with GF at the 1% level, and the first-stage F-statistics far exceed the conventional threshold of 10 ($\Delta$ = [17.05, 78964.97]), ruling out concerns of weak instruments.

**TABLE 5 Results of the first stage regression by instrumental variables.**

| Dependent variable: GF | | | | | | | | | |
|---|---|---|---|---|---|---|---|---|---|
| Instrumental variable | (1) | (2) | (3) | (4) | (5) | (6) | (7) | (8) | (9) |
| L.GF | 0.984*** | | | | | | | | |
| | (258.20) | | | | | | | | |
| L2.GF | | 1.006*** | | | | | | | |
| | | (246.29) | | | | | | | |
| AGF | | | 0.979*** | | | | | | |
| | | | (281.01) | | | | | | |
| L.AGF | | | | 1.000*** | | | | | |
| | | | | (268.77) | | | | | |
| L2.AGF | | | | | 1.022*** | | | | |
| | | | | | (256.91) | | | | |
| GEA | | | | | | 0.140*** | | | |
| | | | | | | (19.89) | | | |
| L.GEA | | | | | | | 0.127*** | | |
| | | | | | | | (17.66) | | |
| L2.GEA | | | | | | | | 0.118*** | |
| | | | | | | | | (15.70) | |
| Pilot | | | | | | | | | 5.898*** |
| | | | | | | | | | (4.13) |
| Control variables | Yes | Yes | Yes | Yes | Yes | Yes | Yes | Yes | Yes |
| F-test | 66666.49 | 60658.25 | 78964.97 | 72237.92 | 66002.29 | 395.79 | 311.77 | 246.47 | 17.05 |
| P-value | 0.0000 | 0.0000 | 0.0000 | 0.0000 | 0.0000 | 0.0000 | 0.0000 | 0.0000 | 0.0000 |
| N | 5767 | 5512 | 5925 | 5679 | 5428 | 6017 | 5767 | 5512 | 6017 |

Notes: IV refers to Instrumental variable; t statistics in parentheses; ***, **, and * denote statistical significance at the 1%, 5%, and 10% levels, respectively.

Based on the second-stage results of the 2SLS regression shown in Table 6, we find that green finance (GF) significantly reduces carbon intensity across all model specifications, regardless of the instrument used. The coefficients remain consistently negative and statistically significant at the 1% level (except for Pilot), confirming the robustness of the baseline findings. Notably, the estimated effects are stronger when using governmental environmental attention (GEA) and its lagged terms (L.GEA, L2.GEA) as instruments. This suggests that cities where environmental issues are more strongly emphasized in political discourse tend to experience greater green finance-driven decarbonization effects.

**TABLE 6 Results of the second stage regression by instrumental variables.**

| Dependent variable: CI | | | | | | | | | |
|---|---|---|---|---|---|---|---|---|---|
| | (1) | (2) | (3) | (4) | (5) | (6) | (7) | (8) | (9) |
| IV | L.GF | L2.GF | AGF | L.AGF | L2.AGF | GEA | L.GEA | L2.GEA | Pilot |
| GF | -0.063*** | -0.054*** | -0.068*** | -0.061*** | -0.053*** | -0.237*** | -0.235*** | -0.232*** | -0.219** |
| | (-16.47) | (-14.45) | (-16.91) | (-15.31) | (-13.42) | (-15.43) | (-13.81) | (-12.16) | (-3.52) |
| IS | 0.013*** | 0.013*** | 0.016*** | 0.015*** | 0.015*** | 0.004 | 0.001 | -0.003 | 0.005 |
| | (3.77) | (3.93) | (4.53) | (4.48) | (4.63) | (0.87) | (0.14) | (-0.35) | (0.85) |
| ES | 1.045*** | 1.049*** | 1.049*** | 1.024*** | 1.032*** | 0.012 | -0.043 | -0.060 | 0.124 |
| | (7.45) | (7.61) | (7.20) | (7.23) | (7.40) | (0.09) | (-0.30) | (-0.39) | (0.30) |
| GT | 0.000*** | 0.000*** | 0.000*** | 0.000*** | 0.000** | 0.000*** | 0.000*** | 0.000*** | 0.000*** |
| | (3.32) | (2.85) | (3.05) | (2.72) | (2.27) | (7.80) | (7.49) | (7.12) | (3.05) |
| EG | 1.722*** | 2.178*** | 1.135** | 1.724*** | 2.178*** | -2.666*** | -2.545*** | -2.461*** | -2.266 |
| | (3.84) | (4.78) | (2.54) | (3.78) | (4.73) | (-4.60) | (-4.08) | (-3.71) | (-1.53) |
| FDI | -0.003*** | -0.003*** | -0.006*** | -0.005*** | -0.005*** | -0.002*** | -0.002*** | -0.002*** | -0.002*** |
| | (-9.75) | (-9.64) | (-13.94) | (-13.86) | (-13.73) | (-5.97) | (-6.13) | (-6.32) | (-5.20) |
| PD | -0.000*** | -0.000*** | -0.000*** | -0.000*** | -0.000*** | 0.000 | 0.000 | 0.000 | -0.000 |
| | (-3.17) | (-3.13) | (-3.20) | (-3.14) | (-3.06) | (0.18) | (0.78) | (1.33) | (-0.08) |
| CONS | 3.249*** | 2.869*** | 3.488*** | 3.186*** | 2.802*** | 9.730*** | 9.840*** | 9.863*** | 9.080*** |
| | (13.37) | (11.70) | (13.95) | (12.73) | (11.15) | (15.73) | (14.06) | (12.35) | (3.91) |
| N | 5767 | 5512 | 5925 | 5679 | 5428 | 6017 | 5767 | 5512 | 6017 |

Notes: IV refers to Instrumental variable; t statistics in parentheses; ***, **, and * denote statistical significance at the 1%, 5%, and 10% levels, respectively.

The Kleiberen Paap test and Cragg Donald test in Table 7 show that the LM statistic and F statistic of all models significantly pass the threshold test, ruling out the possibility of insufficient recognition and weak instrumental variables. Therefore, the estimation results of the two-stage least squares (2SLS) method in this article are reliable.

**TABLE 7 Results of instrumental variables' validation tests.**

| Under-identification test | | | | | | | | | |
|---|---|---|---|---|---|---|---|---|---|
| IV | L.GF | L2.GF | AGF | L.AGF | L2.AGF | GEA | L.GEA | L2.GEA | Pilot |
| Kleibergen-Paap rk LM | 2015.42 | 1871.75 | 2155.26 | 2002.34 | 1841.71 | 373.39 | 296.64 | 235.84 | 12.43 |
| Weak instrumental variable test | | | | | | | | | |
| IV | L.GF | L2.GF | AGF | L.AGF | L2.AGF | GEA | L.GEA | L2.GEA | Pilot |
| Cragg-Donald Wald F | 69884.32 | 64759.76 | 73185.28 | 66508.45 | 59854.54 | 556.91 | 418.78 | 325.41 | 17.13 |
| Stock-Yogo (10%) | 16.38 | 16.38 | 16.38 | 16.38 | 16.38 | 16.38 | 16.38 | 16.38 | 16.38 |

### 4.1.3 Robustness test results

To ensure the robustness of the baseline findings, a series of sensitivity checks are conducted from the following perspectives: (1) excluding municipalities and other special administrative regions to mitigate the potential heterogeneity arising from differences in fiscal capacity and administrative autonomy; (2) applying a 1% winsorization to all variables to minimize the influence of extreme values; (3) re-estimating the model using clustered standard errors to address potential serial correlation and heteroskedasticity within cities.

In addition, Columns (4) further test robustness by replacing the dependent variable with EI (energy consumption intensity), to verify that the observed effects are not driven by the specific selection of emission intensity as the outcome variable. It also conducts a replacement of the core explanatory variable, substituting the green finance (GF) index with representative indicators such as GFN[2] (green financing). These alternative specifications help mitigate concerns about index construction bias and indicator selection.

The robustness test results are reported in Table 8. Across all alternative specifications and sample treatments, the coefficient of green finance (GF) remains significantly negative. This indicates that the decarbonization effect of green finance does not hinge on specific sample choices or model assumptions, thereby confirming the robustness of the main conclusion.

**TABLE 8 Results of robustness test.**

| | (1) | (2) | (3) | (4) | (5) |
|---|---|---|---|---|---|
| GF | -0.1577*** | -0.1462*** | -0.1575** | -0.0589*** | |
| | (-39.572) | (-35.494) | (-15.074) | (-52.751) | |
| GFN | | | | | -0.7733*** |
| | | | | | (-31.939) |
| IS | -0.0357*** | -0.0267*** | -0.0363*** | -0.0116*** | -0.0322*** |
| | (-10.173) | (-8.1868) | (-4.7027) | (-10.751) | (-8.6377) |
| ES | -1.1195*** | -1.4209*** | -1.1131*** | -0.4981*** | -1.1665*** |
| | (-9.6656) | (-5.6919) | (-6.2435) | (-11.878) | (-10.073) |
| GT | 0.0002*** | 0.0003*** | 0.0002*** | 0.0001*** | 0.0001* |
| | (7.6828) | (5.4922) | (3.9866) | (8.0893) | (1.9368) |
| EG | 1.3453*** | 1.8786*** | 1.3540*** | 0.6911*** | 2.7121*** |
| | (5.3421) | (7.4828) | (4.3622) | (8.0863) | (10.239) |
| FDI | -0.0017*** | -0.0039*** | -0.0011** | -0.0007*** | -0.0020*** |
| | (-5.0001) | (-6.7576) | (-1.9710) | (-8.6890) | (-6.9077) |
| PD | -0.0002 | -0.0021*** | -0.0002 | -0.0001 | -0.0013*** |
| | (-1.0087) | (-2.8047) | (-0.2878) | (-1.4314) | (-3.7122) |

Notes: t statistics in parentheses; ***, **, and * denote statistical significance at the 1%, 5%, and 10% levels, respectively.

#### 4.1.4 Spatial econometric model results

To further validate the robustness of the baseline findings, this study examines whether the decarbonization effects of green finance (GF) remain consistent after accounting for potential spatial dependence across cities. Since cities are embedded within broader economic and environmental networks, ignoring spatial correlation may lead to biased estimates[48]. Therefore, we estimate spatial econometric models to test whether the effects of GF extend beyond local jurisdictions and exhibit cross-city spillover patterns. Specifically, we construct an economically weighted geographic distance matrix to capture both spatial proximity and inter-regional economic linkages and re-estimate the benchmark specification accordingly. Table 9 reports the results of the baseline spatial model and two robustness checks. Column (1) presents the baseline spatial lag model, while Columns (2) and (3) report results using an economic-weighted spatial matrix as an alternative to the baseline matrix, and after excluding direct-administered municipalities, respectively.

[2] Green financing (GFN) is proxied by a composite index calculated as the average of standardized values of green credit, green bonds, and green funds.

**TABLE 9 Results of spatial econometric regression.**

| | (1) | | (2) | | (3) | |
|---|---|---|---|---|---|---|
| | Spillover effect | Direct effect | Spillover effect | Direct effect | Spillover effect | Direct effect |
| GF | -0.1411*** | -0.0103 | -0.1397*** | -0.0184 | -0.1363*** | -0.0112 |
| | (-7.747) | (-0.748) | (-7.783) | (-1.454) | (-7.482) | (-0.801) |
| IS | -0.0567** | 0.0497*** | -0.0607*** | 0.0475*** | -0.0550** | 0.0494*** |
| | (-2.367) | (3.163) | (-2.712) | (3.243) | (-2.274) | (2.990) |
| ES | 0.0634 | 1.5448*** | -0.8425 | 1.5706*** | 0.0919 | 1.5180*** |
| | (0.074) | (5.775) | (-0.998) | (5.866) | (0.104) | (5.572) |
| GT | 0.0004*** | 0.0002*** | 0.0004** | 0.0002*** | 0.0004** | 0.0002* |
| | (2.940) | (2.708) | (2.577) | (3.054) | (2.383) | (1.827) |
| EG | 0.4838 | -0.1774 | 0.9168 | -0.2332 | 0.3184 | -0.1368 |
| | (0.503) | (-0.363) | (0.904) | (-0.513) | (0.332) | (-0.276) |
| FDI | -0.0058** | -0.0008 | -0.0069*** | -0.0012 | -0.0078*** | -0.0024*** |
| | (-2.581) | (-0.849) | (-2.665) | (-1.251) | (-2.853) | (-2.580) |
| PD | -0.0002 | -0.0007* | 0.0006 | -0.0009 | -0.0001 | -0.0008* |
| | (-0.212) | (-1.677) | (0.500) | (-2.041) | (-0.097) | (-1.725) |

Notes: z statistics in parentheses; ***, **, and * denote statistical significance at the 1%, 5%, and 10% levels, respectively.

The results confirm that the spatial spillover effect of GF is significantly negative in all model specifications. In the baseline model (Column 1), the spillover coefficient is -0.1411 ($p < 0.01$), suggesting that green finance exerts strong decarbonization effects not only locally, but also in neighboring cities. This supports the existence of spatial transmission mechanisms such as policy imitation, green project diffusion, and inter-city capital flows.

The estimated spillover effect of green finance (GF) remains statistically significant and directionally consistent across alternative model specifications. When the spatial weight matrix is redefined based on inter-city economic linkages (Column 2), the spillover coefficient remains significantly negative ($\beta_{spillover}$= -0.1397, p<0.01). Similarly, when direct-controlled municipalities are excluded from the sample (Column 3), the negative spillover effect persists ($\beta_{spillover}$= -0.1363, p<0.01). These findings provide robust evidence that the estimated effects of GF are not driven by model specification or sample selection and are stable even when accounting for spatial dependence.

**4.1.5 Heterogeneity effects analysis**

To further examine the applicability and effectiveness of green finance across cities with varying development levels, the 285 prefecture-level cities are categorized into five tiers: First-Tier (including new First-Tier), Second-Tier, Third-Tier, Fourth-Tier, and Fifth-Tier cities, in line with China's widely used urban classification. A heterogeneity regression analysis is then conducted for each group.

As shown in Table 10, the estimated coefficients on green finance (GF) are significantly negative across all city tiers at the 1% level, confirming its consistent decarbonization effect nationwide. However, notable differences emerge in the magnitude of the effects. Specifically, green finance exhibits the strongest impact in Fourth-Tier (-0.1472) and Fifth-Tier cities (-0.1344), followed by Third-Tier cities (-0.1134). In contrast, the effects are relatively weaker in First-Tier (-0.1033) and Second-Tier (-0.0841) cities. These results suggest that green finance plays a more prominent role in less-developed cities, potentially due to their greater reliance on external financing and higher marginal returns from green capital.

Overall, the carbon decarbonization effect of green finance is not homogeneous across development levels, highlighting that the effectiveness of green finance depends heavily on the quality of institutional environments, the maturity of financial markets, and local governments' capacity for policy implementation and resource coordination. These findings underscore the institutional adaptability and conditional effectiveness of green finance in city decarbonization.

**TABLE 10 Results of heterogeneity analysis.**

| | Dependent variable: CI | | | | |
|---|---|---|---|---|---|
| | First Tier | Second Tier | Third Tier | Forth Tier | Fifth Tier |
| GF | -0.1033*** | -0.0841*** | -0.1134*** | -0.1472*** | -0.1344*** |
| | (-12.979) | (-7.8057) | (-21.837) | (-17.084) | (-11.780) |
| IS | -0.0158* | 0.0034 | -0.0216*** | -0.0402*** | -0.0219*** |
| | (-1.8109) | (0.2752) | (-4.1339) | (-6.9109) | (-3.2633) |
| ES | 0.0847 | -0.5341*** | -2.6998*** | -1.0732*** | -1.2524*** |
| | (0.2821) | (-3.7092) | (-7.0757) | (-7.6300) | (-5.0817) |
| GT | 0.0000** | 0.0001 | -0.0004*** | -0.0027*** | -0.0051*** |
| | (2.1701) | (0.7940) | (-2.6249) | (-4.5684) | (-2.6092) |
| EG | 0.6014 | 0.3881 | 1.2712*** | 0.9001** | 0.6432 |
| | (1.1373) | (0.7483) | (3.7451) | (2.2050) | (1.3071) |
| FDI | -0.0002 | -0.0008* | -0.0091*** | -0.0150*** | -0.0324*** |
| | (-1.3875) | (-1.7661) | (-10.061) | (-6.2664) | (-4.6599) |
| PD | 0.0009*** | -0.0032** | -0.0011*** | 0.0067*** | -0.0538*** |
| | (5.3967) | (-2.4944) | (-5.6371) | (3.9648) | (-7.5537) |
| N | 432 | 671 | 1525 | 1823 | 1566 |

Notes: t statistics in parentheses; ***, **, and * denote statistical significance at the 1%, 5%, and 10% levels, respectively.

### 4.1.6 Mediation effect analysis

This section systematically investigates the direct mechanisms through which green finance affects urban carbon emission intensity, with a particular focus on the mediating role of key transmission channels such as technology innovation, industrial structure, energy structure, and foreign direct investment. Drawing upon the potential outcomes' framework developed by Imai, Keele [63], this study estimates the Average Causal Mediation Effect (ACME), Average Direct Effect (ADE), and Total Effect, together with 95% confidence intervals. Identifying statistically significant mediation effects - defined as those for which the ACME confidence interval excludes zero - helps to uncover the specific pathways through which green finance operates, thereby making the chain of causal evidence more complete and policy-relevant.

As reported in Column (1), the total effect of green finance on urban carbon emission intensity is significantly negative. The estimated ACME of technological innovation is -0.0015, with a 95% confidence interval of [-0.0029, -0.0002], suggesting that green finance significantly reduces carbon emission intensity through the promotion of technological innovation. The ADE is -0.0613, and the total effect is -0.0628, implying that approximately 2.32% of the total effect is mediated via this channel. This result reflects the catalytic role of green finance in supporting low-carbon R&D, accelerating the diffusion of green technologies, and fostering industrial energy efficiency. However, the relatively modest mediation share also indicates that innovation-based decarbonization tends to be a gradual, long-term process, requiring sustained financial and policy support.

Column (2) demonstrates a more pronounced mediation effect through industrial structure adjustment. The ACME is estimated at -0.0048 and is statistically significant at the 5% level, with a confidence interval of [-0.0061, -0.0036]. The corresponding ADE is -0.0580, and the total effect is -0.0628, suggesting that 7.68% of the total effect of green finance on carbon emission intensity operates through changes in the industrial composition. These results indicate that green finance may facilitate structural transformation by curbing credit to highly polluting sectors and channeling resources toward low-emission industries. Nevertheless, the partial nature of this mediation also implies that market-driven financial signals may not be sufficient for deep industrial restructuring, which may necessitate complementary regulatory or fiscal interventions to accelerate the transition away from carbon-intensive production modes.

Among the four mediators, energy structure exhibits the most substantial mediation effect. As shown in Column (3), the ACME is -0.0119 with a 95% confidence interval of [-0.0139, -0.0099], statistically significant at the 1% level. The total effect of green finance is -0.0628, and the ADE is -0.0509, implying that approximately 18.95% of the total effect is transmitted through energy structure optimization. This finding suggests that green finance plays a pivotal role in reshaping urban energy systems - specifically by promoting clean energy adoption, supporting electrification initiatives, and phasing out fossil-fuel dependency. The relatively high mediated proportion highlights the compatibility between green financial incentives and national energy transition strategies aimed at carbon neutrality. However, the effectiveness of this channel may hinge on regional energy mixes and the responsiveness of local governments to financial policy signals.

Column (4) presents the mediation results through the FDI channel. The ACME is -0.0039 and statistically significant, with a 95% confidence interval of [-0.0052, -0.0027]. The ADE and total effect are -0.0620 and -0.0659, respectively, indicating that 5.86% of the impact of green finance on urban carbon emission intensity is mediated through changes in foreign investment. These results suggest that green finance indirectly shapes urban carbon outcomes by influencing the type and environmental quality of inward foreign capital. Enhanced access to environmentally screened FDI can facilitate the transfer of clean production processes and green technologies. Nonetheless, the relatively limited mediation share underscores the need for more refined environmental standards and screening mechanisms in foreign capital markets to fully harness this pathway's mitigation potential.

**TABLE 11 Mediation effect analysis results.**

| | Mediating Variable: |
|---|---|

| | Technology innovation | Industrial structure | Energy Structure | FDI |
|---|---|---|---|---|
| | (1) | (2) | (3) | (4) |
| ACME | -0.0015<br>(-0.0029, -0.0002) | -0.0048<br>(-0.0061, -0.0036) | -0.0119<br>(-0.0139, -0.0099) | -0.0039<br>(-0.0052, -0.0027) |
| ADE | -0.0613<br>(-0.0677, -0.0548) | -0.0580<br>(-0.0642, -0.0515) | -0.0509<br>(-0.0573, -0.0443) | -0.0620<br>(-0.0685, -0.0554) |
| Total effect | -0.0628<br>(-0.0691, -0.0560) | -0.0628<br>(-0.0699, -0.0556) | -0.0628<br>(-0.0704, -0.0553) | -0.0659<br>(-0.0719, -0.0596) |
| Percentage mediated | 2.32 | 7.68 | 18.95 | 5.86 |

Notes: ACME, ADE, and TOTAL denote average causal mediation effect, average direct effect, and total effect, respectively; estimation results (standard deviation effects) and 95% confidence intervals (in backets) are obtained based on Imai et al. (2010) and Tingley et al. (2015); the number of simulations is 1000.

## 4.2 Nonlinear and marginal effects analysis

### 4.2.1 The marginal and structural effects of green finance

To further identify the most influential drivers of urban carbon emission intensity and quantify the relative contribution of green finance in comparison with other key variables, we employed SHapley Additive exPlanations (SHAP) based on the XGBoost model. Fig.1 presents a combined SHAP summary plot, where the bottom axis shows the distribution of Shapley values (feature contributions to individual predictions), and the top axis illustrates the mean absolute Shapley values (feature importance across the dataset). The color denotes the original feature value (blue: low, red: high)

The SHAP summary plot indicates that green finance (GF) exerts a generally negative, yet somewhat heterogeneous influence on urban carbon emission intensity. As shown in Fig.1a, most Shapley values associated with GF fall below zero, suggesting that in the majority of cities, higher levels of green financial development contribute to reductions in carbon intensity. However, the presence of some positive SHAP values implies that the effect is not uniform across all samples, potentially reflecting regional differences in green finance implementation or effectiveness. Compared to other variables, such as green technology (GT), population density (PD), and energy structure (ES), which show larger mean Shapley values and broader spreads, GF ranks mid-level in overall importance but retains a relatively consistent directional influence. These results underscore the potential of green finance as a policy tool for promoting decarbonization, albeit with varying impacts depending on local contexts.

Building on the preceding analysis, this study further disentangles the heterogeneous effects of green finance, we decomposed the composite indicator into its subcomponents - green bond, green fund, green credit, green equity, green insurance, green investment, and green support - and reassessed their individual contributions using the SHAP framework. As illustrated in Fig.1b, most green finance instruments exhibit non-negligible, albeit differentiated, influences on the predicted carbon intensity. Among them, green bond, green fund, and green credit display relatively higher feature importance, suggesting that these market-oriented financial tools play a more prominent role in shaping carbon outcomes. In contrast, components such as green equity and government green support show weaker and more dispersed effects, indicating limited or context-specific influence. Notably, the SHAP distributions reveal that green bond is predominantly associated with negative Shapley values, highlighting their mitigating impact on urban carbon intensity. This aligns with their theoretical function in channeling financial resources toward low-carbon projects. Conversely, the effect of green insurance and green investment appears more heterogeneous, with Shapley values distributed on both sides of zero, reflecting varying levels of implementation quality or regional institutional differences. Overall, the disaggregated analysis reinforces the importance of distinguishing among policy instruments within the green finance framework to identify the most effective levers for promoting decarbonization.

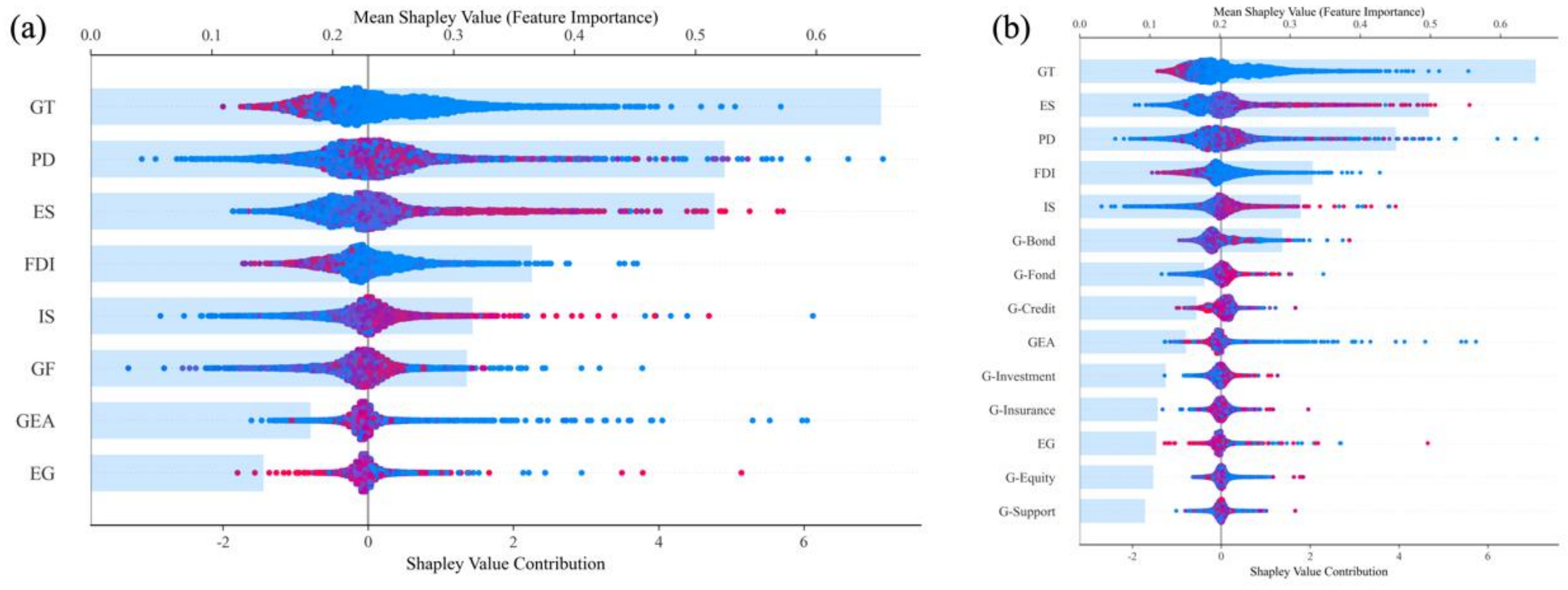

**FIGURE 1**

SHAP-based interpretation of factor contributions to urban carbon emission intensity

### 4.2.2 Heterogeneous decarbonization effects of green finance

To better understand the structural conditions for green finance (GF) to play a stronger or weaker decarbonization role, this section further explore whether the decarbonization effect of green finance (GF) exhibits structural dependence, meaning that its marginal impact on carbon intensity (CI) may vary under different regional development conditions. Specifically, we examine how the effectiveness of GF is conditioned by three key structural variables: energy structure (ES), industrial structure (IS), and green technological capacity (GT). To capture such nonlinear conditional effects, we employ non-parametric estimation methods that allow the marginal effect of GF to vary flexibly across the distribution of each moderator. This approach enables us to identify potential thresholds, inverted-U patterns, or diminishing returns, offering a more nuanced understanding of the conditions under which green finance is most effective in promoting low-carbon development.

Fig.2a illustrates the estimated decarbonization effects of green finance on carbon intensity across Chinese prefecture-level cities in 2022. Spatially, the results reveal a distinct “Core-Periphery” pattern with strong regional clustering. Cities in the Bohai Rim, Yangtze River Delta, Pearl River Delta, and Central China display stronger decarbonization effects, indicating significant advantages in institutional enforcement, project capacity, and financial resource allocation. In contrast, parts of Western, Southwestern, and Northeastern China show weak or absent decarbonization effects, suggesting a considerable gap in implementation capacity and institutional compatibility for green finance in these regions.

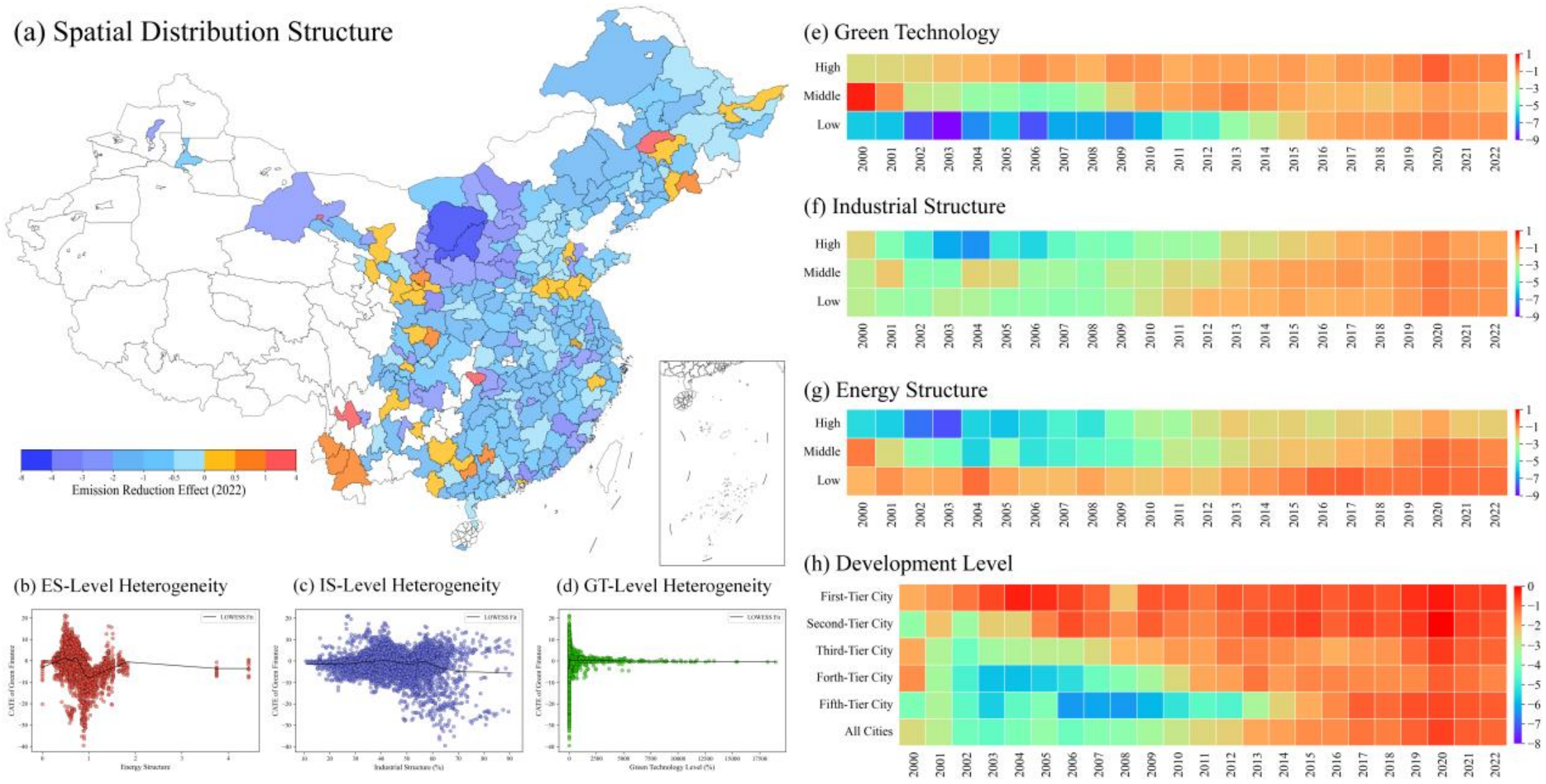

**FIGURE 2**

Spatial distribution, structural heterogeneity, and temporal dynamics of green finance’s decarbonization effects.

Furthermore, by integrating three key conditional variables (ES，IS and GT), this paper examines the marginal effect curve of green finance using non-parametric methods (Fig.2b-d). The results indicate pronounced nonlinearities. For energy structure, the relationship between green finance and decarbonization follows an inverted U-shaped pattern, meaning that policy effectiveness peaks under moderate carbon energy dependence but declines in high-carbon contexts due to limited project scope and energy lock-in effects. Regarding industrial structure, the decarbonization effect of green finance gradually weakens as the share of secondary industry increases, reflecting the lower adaptability of high-carbon industrial systems to green financial instruments. In terms of green technological, marginal effects decrease sharply as innovation capacity improves. Specifically, while effects remain stable at lower technology levels, they decline markedly once the green innovation index exceeds a threshold of 40, suggesting that green finance has greater leverage in low-tech cities but is partially substituted by technological pathways in more advanced regions.

Additionally, a temporal analysis of decarbonization effects from 2000 to 2022 reveals distinct stage-specific patterns in the effectiveness of green finance under varying structural conditions. As shown in Fig.2e, cities with lower levels of GT consistently exhibit stronger policy responsiveness throughout the entire period. In highly industrialized cities and high carbon energy structures cities, the decarbonization effect of green finance show an initial increase followed by a subsequent decrease, reflecting the limitations of green finance in industrialized cities and regions heavily reliant on traditional energy sources (Fig.2g). In terms of city development levels (Fig.2h), the decarbonization effect of green finance shows a clear gradient differentiation, with the marginal intensity of the decarbonization effect decreasing as the level of city development increases. First-Tier cities exhibit persistently weak decarbonization effects across the entire period, with minimal variation in color gradients, suggesting limited overall impact and the absence of a strong green finance-driven transformation. By contrast, Fourth- and Fifth-Tier cities demonstrate the strongest effects during 2002-2011, which then decline over time. This pattern likely reflects the rapid early-stage implementation of national policies in smaller cities through administrative enforcement and resource centralization. However, due to a lack of institutional continuity and sustainable green finance mechanisms, these effects diminished in later years, exhibiting clear signs of marginal decline.

In summary, the carbon decarbonization effects of green finance exhibit strong nonlinearities. Their effectiveness is highly contingent upon a city’s energy mix, industrial base, technological capacity, and stage of development. In structurally weaker or institutionally less developed areas, green finance delivers stronger marginal effects, underscoring its greater leverage as a transitional policy instrument in low-capacity contexts.

## 5. Conclusion and policy recommendations

### 5.1 Conclusion

This study constructs a comprehensive panel dataset of 285 prefecture-level cities in China covering the period from 2000 to 2022, and empirically evaluates the impact of green finance on city carbon intensity. By integrating econometric estimation with non-parametric diagnostic techniques, the analysis provides robust evidence that green finance significantly reduces city carbon intensity, with its effects exhibiting spatial spillovers, conditional heterogeneity, and marginal nonlinear effect, emphasizing the importance of context-specific policy design in advancing low-carbon city transitions. The main findings are summarized as follows:

(1) Green finance significantly reduces city carbon intensity at the aggregate level, with Green Bonds and Green Support exhibiting the strongest decarbonization impacts among various instruments. Spatial analyses further reveal significant spillover effects, whereby green finance contributes to emission reductions both locally and in adjacent cities.

(2) The decarbonization effects of green finance exhibit clear heterogeneity across city tiers. The effects are strongest in Fourth- and Fifth-Tier cities, followed by Third-Tier cities, and are relatively weaker in First- and Second-Tier cities. This gradient highlights the differentiated impact of green finance across urban development levels.

(3) The decarbonization impact of green finance is partially mediated through structural channels, with energy structure optimization exhibiting the strongest effect, followed by industrial upgrading, foreign direct investment, and technological innovation.

(4) The effectiveness of green finance instruments varies considerably across types. Green bonds, green funds, and green credit demonstrate the most substantial decarbonization impacts, whereas instruments such as green equity, insurance, and investment show weaker or more heterogeneous effects, underscoring the differentiated role of financial tools within the green finance framework.

(5) The effectiveness of green finance is reflected in its nonlinear marginal effects across different structural conditions. Its marginal impact is more pronounced in cities with higher carbon dependency, such as those with energy-intensive industries, coal-dominated energy mixes, or weak technological capacity. Moreover, a clear gradient pattern is observed: the marginal effectiveness of green finance tends to diminish as the level of city development increases.

### 5.2 Policy recommendations

Based on the above findings, this study offers the following policy recommendations to enhance the role of green finance in supporting China's "dual carbon" strategy and promoting synergetic city carbon reduction:

First, strengthen the core role of policy-oriented green financial instruments in guiding resource allocation. Given the current immaturity of market-based green finance mechanisms, the government should increase its deployment of fiscal green finance tools, such as dedicated funds, risk compensation, and interest subsidies, to steer financial resources toward key areas including energy conservation, clean energy, and green manufacturing. In parallel, the fiscal financial instrument system should be improved by advancing green bonds, green credit, and green subsidies, alongside the establishment of unified standards and information disclosure mechanisms to enhance transparency, standardization, and sustainability.

Second, implement a tier-specific green finance support strategy across cities. First-Tier cities should prioritize innovation in green financial instruments and integrated institutional applications. Third- and Fourth-Tier cities should focus on strengthening the supply of green projects and the capacity of financial services to improve capital efficiency. Fifth-Tier cities should address foundational institutional weaknesses and project shortages by enhancing fiscal support and administrative capacity to avoid policy inaction or symbolic implementation.

Third, establish cross-regional green finance coordination mechanisms to amplify spatial spillover effects. Existing city agglomerations and regional cooperation platforms should be leveraged to build inter-city green finance coordination networks. Initiatives such as regional green industry funds, joint project development, and information-sharing platforms should be encouraged. In major city clusters such as the Yangtze River Delta, Beijing-Tianjin-Hebei, and Chengdu-Chongqing, unified rules and resource allocation mechanisms should be developed to achieve coordinated implementation and shared decarbonization benefits.

Fourth, promote deep integration of green finance with structural transformation policies. Green finance should be embedded into key policy systems such as green technology promotion, energy structure optimization, and industrial upgrading. A comprehensive governance framework is needed to coordinate fiscal, financial, and industrial policy tools. This includes establishing cross-departmental collaboration platforms, aligning green finance resources with pollution and emission reduction goals, and improving coordination across project preparation, approval, and financing. Supporting policies, joint performance evaluations, and dynamic assessment mechanisms should be introduced to enhance green finance's capacity to support industrial decarbonization, scale up key technologies, and accelerate clean energy investment, thus shifting its function from passive financing support to proactive structural guidance.